\documentclass[10pt,twocolumn,letterpaper]{article}

\usepackage[pagenumbers]{wacv} 

\definecolor{wacvblue}{rgb}{0.21,0.49,0.74}
\usepackage[pagebackref,breaklinks,colorlinks,allcolors=wacvblue]{hyperref}

\usepackage{algorithm}
\usepackage{algpseudocode}
\usepackage{amsmath}
\usepackage{amssymb}

\usepackage{booktabs}
\usepackage{multirow}
\usepackage{comment}

\def\wacvPaperID{1388} 
\def\confName{WACV}
\def\confYear{2026}

\title{Erasing CLIP Memories: Non-Destructive, Data-Free Zero-Shot class Unlearning in CLIP Models}

\author{Ashish Mishra, Tarun Kumar, Gyanaranjan Nayak, Arpit Shah, Suparna Bhattacharya, Martin Foltin\\
Hewlett Packard Labs\\
{\tt\small \{ashish.mishra,tarun.kumar2,arpit.shah,suparna.bhattacharya,martin.foltin\}@hpe.com}
}

\begin{document}
\maketitle
\begin{abstract}
We introduce a novel, closed‐form approach for selective unlearning in multimodal models, specifically targeting pretrained models such as CLIP. Our method leverages nullspace projection to “erase” the target class information embedded in the final projection layer, without requiring any retraining or the use of images from the forget set. By computing an orthonormal basis for the subspace spanned by target text embeddings and projecting these directions, we dramatically reduce the alignment between image features and undesired classes. Unlike traditional unlearning techniques that rely on iterative fine‐tuning and extensive data curation, our approach is both computationally efficient and surgically precise. This leads to a pronounced drop in zero‐shot performance for the target classes while preserving the overall multimodal knowledge of the model. Our experiments demonstrate that even a partial projection can balance between complete unlearning and retaining useful information, addressing key challenges in model decontamination and privacy preservation.
\end{abstract}
    
\section{Introduction}
\label{sec:intro}
Pretrained multimodal models, such as CLIP \cite{radford2021learning}, have transformed the landscape of vision-language understanding by learning rich, joint representations from massive amounts of image and text data. These models demonstrate impressive zero-shot performance on a wide variety of tasks \cite{qian2024online}, making them invaluable for real-world applications. However, as these models are deployed at scale, the need to selectively remove certain associations—whether due to bias \cite{zhou2024limitations}, outdated information \cite{kravets2024zero}, or privacy concerns \cite{liu2024machine, hine2024supporting}, has become increasingly critical. This process, known as \emph{selective unlearning}, poses significant challenges because it demands the eradication of specific target class information without impairing the model’s overall knowledge and performance.

Existing approaches to unlearning often rely on iterative retraining or fine-tuning using additional data samples from the target class, which come with several limitations. 
\begin{itemize}
    \item \textbf{Data Dependence:} They require access to target class data, which might be sensitive, scarce, or proprietary.
    \item \textbf{Computational Cost:} Iterative retraining is computationally expensive and time-consuming, often leading to extended downtime or increased resource usage.
    \item \textbf{Performance Degradation:} The repeated fine-tuning process can inadvertently affect the model's performance on non-target classes, leading to unintended side effects.
\end{itemize}
Notably, class removal can be interpreted as a form of machine unlearning \cite{unlearning_survey}, where specific training data associated with the target class is eliminated. However, implementing this in CLIP is problematic for two main reasons: (a) the original training data for the class to be removed is unavailable, rendering any retraining efforts for unlearning impractical; and (b) given CLIP’s large number of parameters, fine-tuning would be considerably challenging.

\begin{figure}
    \centering
    \includegraphics[width=\linewidth]{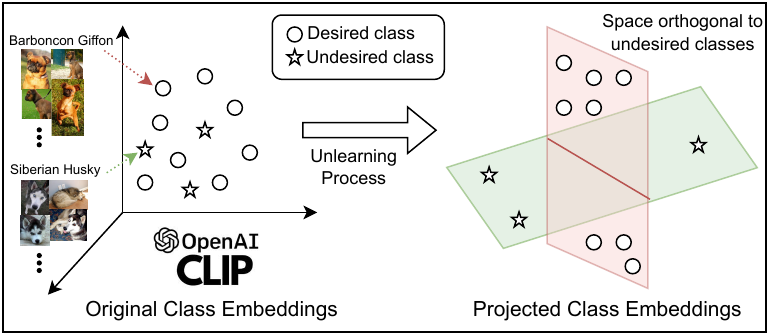}
    \caption{Illustration of projecting original CLIP embeddings to a space where desired classes are projected orthogonal to undesired classes. (The dimensions are for illustration purposes.)}
    \label{fig:intro}
\end{figure}

Motivated by these challenges, we propose a novel, data-free method for selective class unlearning that leverages \emph{nullspace projection}. Our approach, inspired by earlier works on nullspace projection for bias mitigation \cite{ravfogel2020null}, circumvents the need for target/unlearn class samples and the burdens of retraining by directly modifying the final projection layer of the model in a closed-form, non-iterative manner. Specifically, as shown in Figure \ref{fig:intro}, by computing an orthonormal basis for the subspace spanned by target text embeddings and projecting them from the image embeddings, we effectively remove the undesired associations.

Class unlearning has been extensively explored in unimodal deep learning models, but it has received less attention in multimodal frameworks like CLIP, which are inherently more challenging and applicable. To the best of our knowledge, only two studies—ZSCCF and ZSCUC—have addressed zero-shot class unlearning in the CLIP model. In contrast to existing methods—ZSCUC, which relies on synthesized examples, and ZSCCF, which uses a projection method with LoRA weights and LLM-generated fine-grained class text for retention—our proposed Consistent Class Unlearning Projection (CCUP) offers a closed-form solution for obtaining the projection matrix that seamlessly handles identity preservation, forgetting, and retention of non-forgot classes. Moreover, CCUP operates entirely data-free, eliminating the need for any fine-tuning or retraining, which significantly reduces computational overhead and minimizes potential errors in the unlearning process.

The key advantages of our proposed approach are:
\begin{itemize}
    \item \textbf{Data-Free Unlearning:} Our method does not require any additional data from the target (forget) classes, making it particularly suitable for scenarios with data scarcity or privacy concerns.
    \item \textbf{Computational Efficiency:} The closed-form computation via nullspace projection eliminates the need for iterative retraining, thereby significantly reducing computational overhead and speeding up the unlearning process.
    \item \textbf{Selective Preservation:} By surgically removing only the target subspace, our method preserves the model's broader multimodal knowledge, ensuring minimal degradation in performance on non-target classes.
    \item \textbf{Ease of Implementation:} The approach integrates seamlessly with pretrained models such as CLIP, requiring only a modification of the projection matrix, and can be easily adjusted (e.g., via a tuning parameter) to achieve varying degrees of unlearning.
\end{itemize}

The novelty of our work lies in its ability to achieve selective unlearning in a completely data-free setup—a departure from traditional retraining-based methods. By eliminating the dependency on target class samples, our method addresses both computational and privacy challenges inherent in current approaches. Moreover, the application of nullspace projection in multimodal models is an innovative contribution that opens up new research directions in model decontamination and bias mitigation \cite{huo2025mmunlearner}.

In this paper, we present the detailed mathematical formulation, implementation strategy, and evaluation protocols for our nullspace projection unlearning method. We demonstrate through extensive experiments that our approach not only effectively reduces the alignment between target text and non-target images' embeddings but also maintains competitive performance on non-target classes. This balance of efficiency, precision, and data independence marks a significant step forward in the development of robust, privacy-preserving multimodal systems.

\section{Related Work}
Early machine unlearning techniques focused on dataset partitioning and selective retraining of the models \cite{bourtoule2021machine}. Such methods fundamentally fail in multimodal contexts present in models such as CLIP. Diffusion models advanced concept removal through attention manipulation \cite{limachine}. However, these methods risk the generation quality and require significant training compute. Text-guided approaches such as SalUn \cite{fan2023salun} depend on anchor prompts (e.g., "vanilla photo") for style removal, failing for abstract concepts lacking linguistic descriptors.

Recent CLIP-focused methods reveal three dominant strategies: CLIPErase \cite{yang2024ULcliperase} introduces a triple-module framework (Forgetting/Retention/Consistency) that disentangles visual-textual associations but requires access to training pairs for retention control. In a zero-shot setting, \cite{kravets2025ULzeroshotclip0} introduces a projection matrix adjustment method that operates without visual data. However, this approach risks the cross-modal alignment of retained classes due to its exclusive focus on textual embeddings. We use this method as a baseline in our experiments. We use another baseline \cite{kravets2024ULzeroshotclip1} that employs Lipschitz-regularized synthetic data generation to smooth class boundaries. While eliminating real data dependencies, its layer-wise parameter updates introduce catastrophic forgetting risks.

Existing approaches face four constraints: (1) vision-only methods cannot resolve multimodal entanglement in CLIP's joint space, (2) diffusion techniques require generative components incompatible with discriminative models, (3) synthetic-sample methods induce risk of catastrophic forgetting through gradient perturbations, and (4) all current CLIP unlearning strategies necessitate either training data access or intensive retaining. We empirically compare the methods closest to our proposed work \cite{kravets2025ULzeroshotclip0, kravets2024ULzeroshotclip1} in the later sections. Our closed-form nullspace projection eliminates these issues through orthogonal decomposition of text embeddings in CLIP’s final layer

\section{Methodology}

This section outlines our proposed data--free and training--free approach for class unlearning in pretrained CLIP models. Our goal is to remove unwanted class--specific information (the ``forget'' classes) from the joint image--text embedding space while retaining the useful semantic structure for the ``retain'' classes---all without any retraining or fine tuning of the CLIP model.

\subsection{Problem Definition}

Vision--language models like CLIP learn a joint embedding space that encodes rich semantic information. However, when certain classes need to be forgotten (e.g., due to privacy or bias concerns), it is essential to eliminate class--specific information without harming the representation of other classes. The standard null--space projection method achieves this by completely removing the subspace aligned with the forget classes. Yet, in fine--grained domains where the subspaces of forget and retain classes may overlap, a hard projection risks discarding useful features.

Our objective is to design a closed--form linear transformation, \(W \in \mathbb{R}^{d\times d}\), that meets the following criteria:
\begin{enumerate}
    \item \textbf{Suppress Forget--Class Features:} Remove or significantly reduce the components of the forget classes.
    \item \textbf{Preserve Retain--Class Features:} Keep, or even reinforce, the features that represent the retain classes.
    \item \textbf{Minimize Distortion:} Remain close to the identity transformation to maintain the overall structure of the embedding space.
\end{enumerate}

\subsection{Preliminaries and Notation}

Let:
\begin{itemize}
    \item \(d\) be the dimension of the joint embedding space.
    \item \(T_f \in \mathbb{R}^{d \times m_f}\) be the matrix with normalized text embeddings for the \emph{forget} classes (\(m_f\) classes).
    \item \(T_r \in \mathbb{R}^{d \times m_r}\) be the matrix with normalized text embeddings for the \emph{retain} classes (\(m_r\) classes).
    \item \(I \in \mathbb{R}^{d \times d}\) be the identity matrix.
\end{itemize}

The standard null--space projection is given by:
\[
P = I - T_f \left(T_f^T T_f\right)^{-1} T_f^T,
\]
which removes all components of any image feature \(x \in \mathbb{R}^d\) lying in the subspace spanned by \(T_f\). While effective at erasing forget--class information, this approach may remove overlapping features needed for the retain classes.

\begin{figure*}
    \centering    \includegraphics[width=0.85\linewidth]{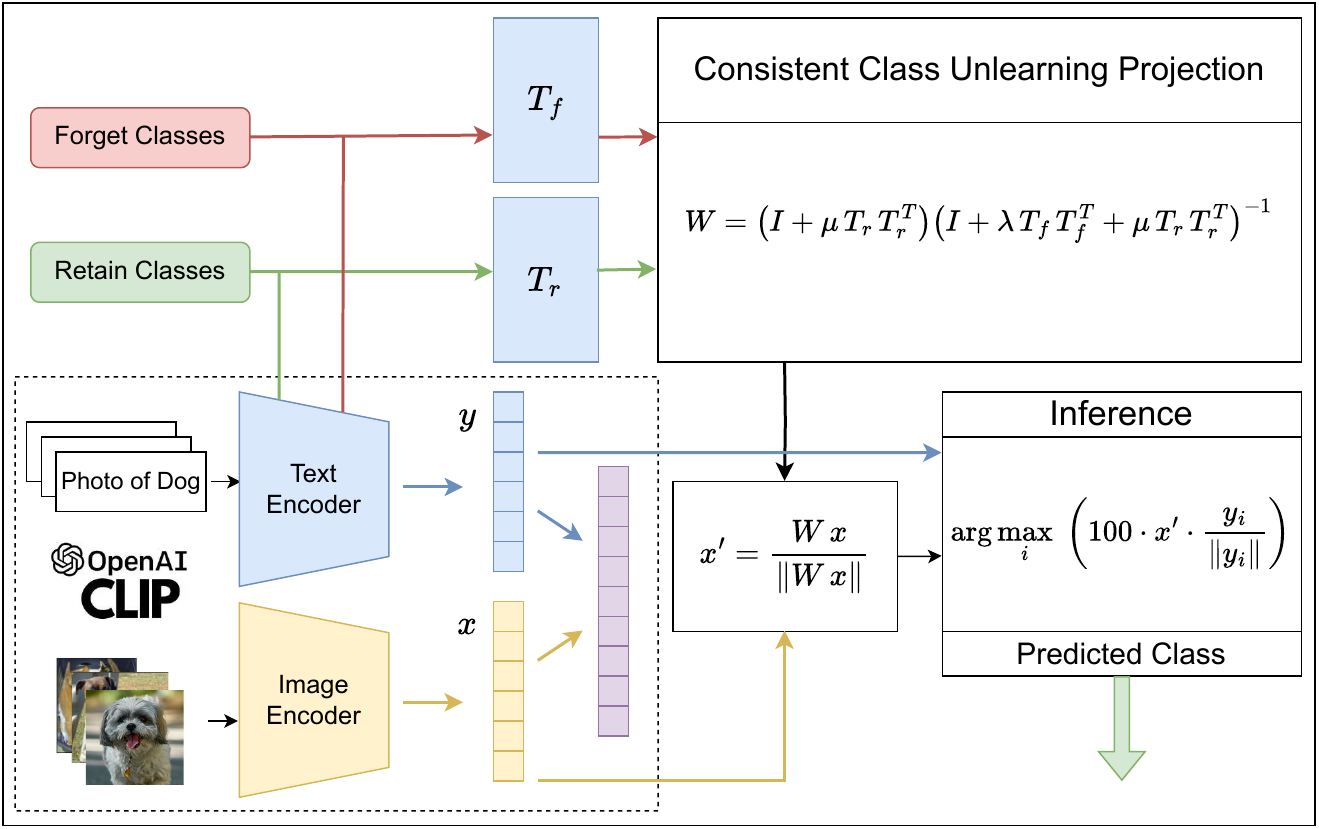}
    \caption{Proposed architecture for unlearning \textit{Forget Classes} using proposed CCUP projection method.}
    \label{fig:method}
\end{figure*}

\subsection{Proposed Consistent Class Unlearning Projection (CCUP)}

To balance unlearning and retention, we propose the \emph{Consistent Class Unlearning Projection} (CCUP). Instead of entirely erasing the forget subspace, CCUP finds a transformation \(W\) that partially suppresses forget--class features while preserving the retain--class structure. We achieve this by optimizing the following objective:
\[
\min_{W} \; \|W - I\|_F^2 \;+\; \lambda\, \|W\,T_f\|_F^2 \;+\; \mu\, \|W\,T_r - T_r\|_F^2,
\]
where:
\begin{itemize}
    \item \(\|W - I\|_F^2\) penalizes deviations from the identity matrix, ensuring minimal overall distortion.
    \item \(\lambda\, \|W\,T_f\|_F^2\) encourages suppression of the forget--class components (with \(\lambda>0\) controlling the strength of this suppression).
    \item \(\mu\, \|W\,T_r - T_r\|_F^2\) enforces that the retain--class features remain largely unaltered (with \(\mu>0\) controlling the degree of retention).
\end{itemize}

Setting the derivative with respect to \(W\) to zero yields the closed--form solution:
\[
W = \left(I + \mu\, T_r\,T_r^T\right) \left(I + \lambda\, T_f\,T_f^T + \mu\, T_r\,T_r^T\right)^{-1}.
\]

\subsubsection{Insight into the Formulation}

\textbf{Identity Preservation:}\\[0.5em]
The term \(\|W - I\|_F^2\) ensures that \(W\) does not stray far from the identity, thereby maintaining the general structure of the CLIP features.

\medskip

\textbf{Forgetting Mechanism:}\\[0.5em]
The term \(\lambda\, \|W\,T_f\|_F^2\) forces the transformation to suppress the components associated with the forget classes. A larger \(\lambda\) enforces a stronger reduction of these unwanted features.

\medskip

\textbf{Retention Mechanism:}\\[0.5em]
The term \(\mu\, \|W\,T_r - T_r\|_F^2\) ensures that features relevant to the retain classes are preserved. Increasing \(\mu\) minimizes the deviation of the retain class features after transformation.

\medskip

\textbf{Balanced Trade--Off:}\\[0.5em]
The closed--form solution,
\[
W = \left(I + \mu\, T_r\,T_r^T\right) \left(I + \lambda\, T_f\,T_f^T + \mu\, T_r\,T_r^T\right)^{-1},
\]
nicely balances these competing objectives. In extreme cases:
\begin{itemize}
    \item When \(\mu=0\) and \(\lambda \to \infty\), the solution approximates the standard null--space projection:
    \[
    W \approx I - T_f \left(T_f^T T_f\right)^{-1} T_f^T.
    \]
    \item When \(\lambda=0\) and \(\mu\) is large, \(W \approx I\), implying that no unlearning occurs.
\end{itemize}
Thus, by adjusting \(\lambda\) and \(\mu\), CCUP selectively removes forget--class information while maintaining the integrity of retain--class features.

\subsection{Evaluation Process}

Once the projection matrix \(W\) is computed, it is applied to any image feature \(x\in\mathbb{R}^d\) as follows:
\[
x' = \frac{W\,x}{\|W\,x\|}.
\]
The transformed feature \(x'\) is then used for zero--shot classification via cosine similarity with the CLIP text embeddings. For a test image, the predicted class is given by:
\[
\text{predicted class} = \arg\max_{i} \; \left(100 \cdot x' \cdot \frac{y_i}{\|y_i\|}\right),
\]
where \(y_i\) is the CLIP text embedding for the \(i^\text{th}\) class. By separately evaluating the performance on the forget and retain subsets, we verify that:
\begin{itemize}
    \item The forget--class accuracy significantly drops (i.e., cosine similarities are near zero).
    \item The retain--class accuracy remains high.
\end{itemize}

\begin{algorithm}[h]
\caption{Consistent Class Unlearning Projection (CCUP)}
\begin{algorithmic}[1]
\Require 
  \begin{itemize}
    \item $T_f \in \mathbb{R}^{d \times m_f}$: Normalized text embeddings for the \emph{forget} classes.
    \item $T_r \in \mathbb{R}^{d \times m_r}$: Normalized text embeddings for the \emph{retain} classes.
    \item Regularization parameters: $\lambda > 0$ and $\mu > 0$.
    \item $I \in \mathbb{R}^{d \times d}$: Identity matrix.
  \end{itemize}
\Ensure Transformation matrix $W \in \mathbb{R}^{d \times d}$.
  
\State Compute $M_f \gets \lambda\, T_f\, T_f^T$.
\State Compute $M_r \gets \mu\, T_r\, T_r^T$.
\State Form denominator: $D \gets I + M_f + M_r$.
\State Form numerator: $N \gets I + M_r$.
\State Compute the projection matrix:
\[
W \gets N \cdot D^{-1} = \left(I + \mu\, T_r\,T_r^T\right) \left(I + \lambda\, T_f\,T_f^T + \mu\, T_r\,T_r^T\right)^{-1}.
\]
\For{each image feature $x \in \mathbb{R}^d$}
    \State Compute transformed feature: $x' \gets W\,x$.
    \State Normalize: $x' \gets \dfrac{x'}{\|x'\|}$.
\EndFor
\For{each class $i$ with CLIP text embedding $y_i$}
    \State Compute similarity: 
    \[
    s_i \gets 100 \cdot \left( x' \cdot \frac{y_i}{\|y_i\|} \right).
    \]
\EndFor
\State \Return Predicted class: $\arg\max_i\, s_i$.
\end{algorithmic}
\end{algorithm}

\section{Experimental setup and Results}

\subsection{Datasets}
We assess CLIP's forgetting capabilities using four high-quality, fine-grained datasets. Caltech101 \cite{caltech101}, features images from 101 unique categories that represent various objects and scenes. StanfordCars \cite{stanfordcars} includes images of cars spanning different makes and models. OxfordFlowers \cite{oxford-102-flower-pytorch} contains images of flowers across 102 distinct classes, and StanfordDogs \cite{KhoslaYaoJayadevaprakashFeiFei_FGVC2011} comprises 120 classes of various dog species. We also conducted evaluations on two predominantly zero-shot learning image datasets: AWA2 \cite{awa2} and CUB \cite{cub-200-birds_dataset}. To demonstrate the scalability and generalizability of our approach, we conducted experiments on both the Tiny ImageNet and Mini-ImageNet datasets.

\subsection{Evaluation Metrics}
Our goal is to minimize the accuracy on the forget classes while keeping the accuracy on the remaining classes at levels comparable to those observed before forgetting. We evaluate our approach by measuring the accuracy of the forget and retain classes both before and after forgetting in the pretrained CLIP model. The BF (before forgetting) and AF (after forgetting) accuracies clearly demonstrate the decrease in performance for the forget classes while maintaining stability for the retain classes. We evaluated the effectiveness of our approach using the Membership Inference Attack (MIA) score. In the context of machine unlearning, the MIA score quantifies how well a model "forgets" information about data designated for removal (the "forget set") while continuing to retain knowledge about the remaining data (the "retain set"). We calculated the MIA as:
\begin{equation}
\text{MIA} = \left( \text{BF}_{\text{forget}} - \text{AF}_{\text{forget}} \right)
           - \left( \text{BF}_{\text{retain}} - \text{AF}_{\text{retain}} \right)
\end{equation}
A low MIA score indicates a more effective unlearning approach, as it shows that the model has successfully forgotten the designated classes while retaining knowledge of the remaining classes.

\begin{table*}[ht]
\centering
\caption{Result Comparison Across Datasets: Unlearning Performance for Forgetting and Retain Classes (ViT-B/32 and ViT-B/16) with MIA scores}
\resizebox{\textwidth}{!}{%
\begin{tabular}{l
                |cc|cc|c
                |cc|cc|c
                |cc|cc|c
                |cc|cc|c}
\toprule
\multirow{4}{*}{\textbf{Method}} 
   & \multicolumn{5}{c|}{\textbf{StanfordCars}} 
   & \multicolumn{5}{c|}{\textbf{StanfordDogs}} 
   & \multicolumn{5}{c|}{\textbf{Caltech101}} 
   & \multicolumn{5}{c}{\textbf{OxfordFlowers}} \\
\cmidrule(lr){2-6} \cmidrule(lr){7-11} \cmidrule(lr){12-16} \cmidrule(lr){17-21}
  & \multicolumn{2}{c|}{\textbf{Forgetting}} & \multicolumn{2}{c|}{\textbf{Retain}} & \multirow{2}{*}{\textbf{MIA}}
  & \multicolumn{2}{c|}{\textbf{Forgetting}} & \multicolumn{2}{c|}{\textbf{Retain}} & \multirow{2}{*}{\textbf{MIA}}
  & \multicolumn{2}{c|}{\textbf{Forgetting}} & \multicolumn{2}{c|}{\textbf{Retain}} & \multirow{2}{*}{\textbf{MIA}}
  & \multicolumn{2}{c|}{\textbf{Forgetting}} & \multicolumn{2}{c|}{\textbf{Retain}} & \multirow{2}{*}{\textbf{MIA}} \\
  \cmidrule(lr){2-3} \cmidrule(lr){4-5}
  \cmidrule(lr){7-8} \cmidrule(lr){9-10}
  \cmidrule(lr){12-13} \cmidrule(lr){14-15}
  \cmidrule(lr){17-18} \cmidrule(lr){19-20}
  & BF & AF & BF & AF & 
  & BF & AF & BF & AF & 
  & BF & AF & BF & AF & 
  & BF & AF & BF & AF &  \\
\midrule
\multicolumn{21}{c}{\textbf{ViT-B/32}} \\
\midrule
Ours     & 82.23 & 0.38 & 81.45 & 75.42 & \textbf{75.82}
         & 80.38 & 0.00 & 60.18 & 50.30 & \textbf{70.50}
         & 83.76 & 0.00 & 95.13 & 94.44 & \textbf{83.07}
         & 71.80 & 0.00 & 70.50 & 69.10 & \textbf{70.40} \\
ZSL-CLIP & 82.20 & 1.82 & 81.40 & 52.30 & 51.28
         & 80.38 & 4.75 & 60.15 & 15.92 & 31.40
         & 83.76 & 1.15 & 95.13 & 70.54 & 58.02
         & 71.80 & 0.08 & 70.50 & 68.40 & 69.62 \\
Lip      & 82.20 & 12.50 & 81.40 & 56.72 & 45.02
         & 80.38 & 16.23 & 60.15 & 23.60 & 27.60
         & 83.76 & 20.50 & 95.13 & 70.50 & 38.63
         & 71.80 & 0.35 & 70.50 & 61.90 & 62.85 \\
Emb      & 82.20 & 10.50 & 81.40 & 66.20 & 56.50
         & 80.38 & 16.23 & 60.15 & 32.60 & 36.60
         & 83.76 & 17.50 & 95.13 & 70.80 & 41.93
         & 71.80 & 1.35 & 70.50 & 63.70 & 63.65 \\
Amns     & 82.20 & 8.55 & 81.40 & 32.70 & 24.95
         & 80.38 & 18.30 & 60.15 & 35.90 & 37.83
         & 83.76 & 18.00 & 95.13 & 69.10 & 39.73
         & 71.80 & 0.75 & 70.50 & 60.30 & 60.85 \\
EMMN     & 82.20 & 14.80 & 81.40 & 36.72 & 22.72
         & 80.38 & 9.80 & 60.15 & 39.50 & 49.93
         & 83.76 & 18.80 & 95.13 & 78.00 & 47.83
         & 71.80 & 0.55 & 70.50 & 62.90 & 63.65 \\
\midrule
\multicolumn{21}{c}{\textbf{ViT-B/16}} \\
\midrule
Ours     & 80.30 & 0.38 & 80.45 & 75.10 & \textbf{74.57}
         & 79.50 & 0.00 & 60.18 & 52.20 & \textbf{71.52}
         & 80.66 & 0.00 & 94.90 & 94.41 & \textbf{80.17}
         & 70.60 & 0.00 & 70.10 & 68.90 & \textbf{69.40} \\
ZSL-CLIP & 80.30 & 1.80 & 80.40 & 52.90 & 51.00
         & 78.60 & 4.92 & 58.15 & 18.22 & 33.75
         & 80.66 & 9.15 & 94.90 & 70.20 & 46.81
         & 70.60 & 0.08 & 70.10 & 62.60 & 63.02 \\
Lip      & 80.30 & 22.20 & 81.40 & 55.82 & 32.52
         & 79.90 & 18.50 & 60.15 & 23.80 & 25.05
         & 83.76 & 21.30 & 95.13 & 70.12 & 37.45
         & 70.60 & 0.29 & 70.10 & 60.50 & 60.71 \\
Emb      & 80.30 & 12.20 & 81.40 & 68.20 & 54.90
         & 79.90 & 15.50 & 60.15 & 33.80 & 38.05
         & 83.76 & 11.30 & 95.13 & 73.10 & 50.43
         & 70.60 & 1.90 & 70.10 & 62.50 & 61.10 \\
Amns     & 80.30 & 10.50 & 81.40 & 27.82 & 16.22
         & 79.90 & 18.10 & 60.15 & 37.80 & 39.45
         & 83.76 & 18.30 & 95.13 & 67.20 & 37.53
         & 70.60 & 0.89 & 70.10 & 59.10 & 58.71 \\
EMMN     & 80.30 & 15.30 & 81.40 & 32.10 & 15.70
         & 79.90 & 10.80 & 60.15 & 43.20 & 52.15
         & 83.76 & 20.10 & 95.13 & 76.50 & 45.03
         & 70.60 & 0.76 & 70.10 & 63.80 & 63.54 \\
\bottomrule
\end{tabular}%
}
\label{tab:tab1}
\end{table*}

\subsection{Dataset Splits}
For the fine-grained datasets—StanfordDogs, StanfordCats, Caltech101, and OxfordFlowers—we split the classes by assigning $40\%$ to the forget set and the remaining $60\%$ to the retain set. In contrast, for the AWA2 and CUB datasets, we adopted a zero-shot setup where unseen classes serve as the forget set and seen classes as the retain set. Specifically, AWA2 and CUB have 10 and 50 classes in the forget set, while the remaining 40 and 150 classes constitute the retain set, respectively. For the Tiny-ImageNet dataset, which contains 200 classes, we divided the classes into 50 for the forget set and 150 for the retain set. Similarly, for the Mini-ImageNet dataset with 100 classes are split into 40 for the forget set and 60 for the retain set.

\subsection{Baseline}
To the best of our knowledge, only a few existing methods—specifically ZSL-CLIP \cite{kravets2025ULzeroshotclip0}, Lip \cite{lip}, Emb, Amns, and EMMN—are suitable for comparison with our proposed approach. Therefore, we use these methods as baselines to evaluate and demonstrate the effectiveness of our approach. Furthermore, to showcase the broader applicability of our method, we applied it to two variants of CLIP: ViT-B/16 and ViT-B/32. Due to space limitation we included some results for CLIP: ViT-B/16 in the supplementary material.  

\begin{table*}[ht]
\centering
\caption{Result Comparison for Two Datasets: Unlearning Performance for Forgetting and Retain Classes (including MIA score)}
\begin{tabular}{l|cccc|c|cccc|c}
\toprule
\multirow{2}{*}{\textbf{Method}} 
   & \multicolumn{4}{c|}{\textbf{CUB}} & \textbf{MIA} 
   & \multicolumn{4}{c|}{\textbf{AWA2}} & \textbf{MIA} \\
\cmidrule(lr){2-5} \cmidrule(lr){7-10}
 & \multicolumn{2}{c|}{Forgetting Classes} & \multicolumn{2}{c|}{Retain Classes} 
 &  
 & \multicolumn{2}{c|}{Forgetting Classes} & \multicolumn{2}{c|}{Retain Classes} 
 & \\
\cmidrule(lr){2-3} \cmidrule(lr){4-5} \cmidrule(lr){7-8} \cmidrule(lr){9-10}
 & BF & AF & BF & AF 
 & 
 & BF & AF & BF & AF 
 & \\
\midrule
Ours      & 40.94 & 0.00 & 38.50 & 25.60 & \textbf{28.04} 
          & 89.64 & 0.00 & 92.49 & 88.32 & \textbf{85.47} \\
ZSL-CLIP  & 41.21 & 3.08 & 34.43 & 18.91 & 22.61 
          & 88.42 & 12.85 & 88.38 & 36.75 & 23.94 \\
Lip       & 41.21 & 8.91 & 34.43 & 10.34 & 8.21 
          & 88.42 & 20.92 & 88.38 & 31.23 & 10.35 \\
Emb       & 41.21 & 5.10 & 34.43 & 13.40 & 15.08 
          & 88.42 & 12.20 & 88.38 & 61.23 & 49.07 \\
Amns      & 41.21 & 7.90 & 34.43 & 16.34 & 15.22 
          & 88.42 & 15.10 & 88.38 & 49.30 & 34.24 \\
EMMN      & 41.21 & 10.10 & 34.43 & 16.12 & 12.80 
          & 88.42 & 17.25 & 88.38 & 42.80 & 25.59 \\
\bottomrule
\end{tabular}%
\label{tab:tab2}
\end{table*}

\begin{table*}[ht]
\centering
\caption{Result Comparison for Two Datasets: Unlearning Performance for Forgetting and Retain Classes (with MIA score)}
\begin{tabular}{l|cccc|c|cccc|c}
\toprule
\multirow{2}{*}{\textbf{Method}} 
   & \multicolumn{4}{c|}{\textbf{Tiny-ImageNet}} & \textbf{MIA} 
   & \multicolumn{4}{c|}{\textbf{Mini-ImageNet}} & \textbf{MIA} \\
\cmidrule(lr){2-5} \cmidrule(lr){7-10}
 & BF & AF & BF & AF &  & BF & AF & BF & AF & \\
\cmidrule(lr){2-3} \cmidrule(lr){4-5} \cmidrule(lr){7-8} \cmidrule(lr){9-10}
 & \multicolumn{2}{c|}{Forgetting Classes} & \multicolumn{2}{c|}{Retain Classes} &  
 & \multicolumn{2}{c|}{Forgetting Classes} & \multicolumn{2}{c|}{Retain Classes} & \\
\midrule
Ours & 55.40 & 9.32 & 53.80 & 48.60 & \textbf{40.88} 
     & 60.50 & 8.25 & 58.7 & 52.23 & \textbf{45.78} \\
ZSL-CLIP & 55.40 & 12.08 & 53.80 & 43.91 & 33.43 
         & 60.50 & 19.5 & 58.7 & 39.52 & 21.82 \\
Lip & 55.40 & 20.11 & 53.80 & 35.34 & 16.83 
    & 60.50 & 17.72 & 58.7 & 38.56 & 22.64 \\
Emb & 55.40 & 14.66 & 53.80 & 38.40 & 25.34 
    & 60.50 & 23.60 & 58.7 & 39.90 & 18.10 \\
Amns & 55.40 & 22.91 & 53.80 & 28.52 & 7.21 
    & 60.50 & 26.20 & 58.7 & 41.63 & 17.23 \\
EMMN & 55.40 & 26.10 & 53.80 & 37.66 & 13.16 
    & 60.50 & 21.50 & 58.7 & 44.55 & 24.85 \\
\bottomrule
\end{tabular}
\label{tab:imgnet}
\end{table*}

\subsection{Result Analysis}
To demonstrate the effectiveness of our proposed approach, we conducted experiments on four fine-grained datasets—StanfordDogs, StanfordCars, Caltech101, and OxfordFlowers, as well as on the datasets used for zero-shot image classification, AWA2, and CUB. Additionally, to highlight the generalizability of our method, we evaluated it on Tiny-ImageNet and Mini-ImageNet. As shown in Table \ref{tab:tab1}, our unlearning approach outperformed all baselines for the fine-grained datasets in terms of both forgetting and retaining classes. This indicates that our method successfully unlearns the designated classes from the pretrained CLIP model without affecting the information of the remaining classes. In other words, our approach effectively handles the unlearning of targeted classes while preserving the knowledge of non-targeted classes in a data-free setup, which supports the validity of our projection space that nullifies the targeted classes and maintains the others in the projected space. A similar observation is evident in Table \ref{tab:tab2} for the zero-shot classification datasets, AWA2 and CUB, which further highlights the generalizability of our proposed projection space for unlearning within the CLIP model. This is achieved without the need for fine-tuning or iterative learning in a data-free framework. Table \ref{tab:imgnet} presents the results for the Tiny-ImageNet and Mini-ImageNet datasets. In these experiments, we fine-tuned the pretrained CLIP model for a few epochs before applying unlearning to ensure that the model was exposed to all relevant classes, avoiding any unseen class scenarios. Our results demonstrate that the proposed unlearning approach outperforms the baselines in both forgetting and retention metrics. Additionally, our method achieves a consistently high MIA score across all datasets, highlighting its effectiveness in removing knowledge of the forget set while preserving information about the retained classes.

We evaluated our proposed approach on two different CLIP versions, ViT-B/16 and ViT-B/32. In both cases, our method demonstrated significant performance in unlearning and preserving capabilities across all datasets, which supports its usability across various model types.

For a fair comparison, we also conducted experiments using the ZSL-CLIP setup, as shown in Table \ref{tab:tab3}. In this configuration, only a small number of classes were designated for forgetting. We observed that our proposed approach achieves comparable performance in both unlearning and retaining accuracies when compared with the baselines, Lip and ZSL-CLIP.

Overall, the results from Tables \ref{tab:tab1}, \ref{tab:tab2}, and \ref{tab:tab3} indicate that our proposed approach consistently performs well across both scenarios—whether dealing with a large number of targeted/forgetting classes or a small subset—in terms of both unlearning and retention accuracy. In contrast, the baseline models, Lip and ZSL-CLIP, perform significantly better when only a few classes are forgotten. However, as the number of forgetting classes increases, while their unlearning accuracy remains competitive, their retention accuracy declines. This may be because unlearning a large number of classes, which semantically overlap with the classes to be retained, disrupts the overall semantic structure in the projection space. To address this issue, we specifically incorporated an identity preservation constraint in our projection approach, which helps maintain the general structure of the CLIP features in the projection space.

\textbf{MIA Score Analysis:}
Our approach consistently achieves a higher MIA score across all datasets compared to existing baselines, clearly demonstrating its superiority in selective unlearning. This high MIA score directly indicates that our method is able to effectively and selectively forget the target (forget) classes while maintaining strong performance on the remaining (retain) classes.

\begin{table*}[ht]
\centering
\caption{Forgetting on multiple classes with RN50 and ViT-B/16 models, plus additional rows for ViT-32 and ViT, with Target Class BF/AF.}
\resizebox{\textwidth}{!}{%
\begin{tabular}{l l l l | cc | cc | cc | cc | cc}
\toprule
\multicolumn{4}{c|}{Basic Info} & \multicolumn{2}{c|}{Target Classes} & \multicolumn{2}{c|}{StanfordCars} & \multicolumn{2}{c|}{StanfordDogs} & \multicolumn{2}{c|}{Caltech101} & \multicolumn{2}{c}{OxfordFlowers} \\
\cmidrule(lr){1-4} \cmidrule(lr){5-6} \cmidrule(lr){7-8} \cmidrule(lr){9-10} \cmidrule(lr){11-12} \cmidrule(lr){13-14}
Method & Model & Dataset & Classes & BF & AF & BF & AF & BF & AF & BF & AF & BF & AF \\
\midrule
Lip      & RN50     & StanfordDogs   & Pekinese, toy poodle, Scotch terrier           & 0.591 & 0.091 & 0.558 & 0.547 & --    & --    & 0.857 & 0.865 & 0.661 &  0.633\\
Lip      & RN50     & StanfordCars   & 2009 Spyker C8 Coupe, 2010 Dodge Ram Pickup 3500 Crew Cab, 2011 Ford Ranger SuperCab  & 0.397 & 0.222 & --    & --    & 0.517 & 0.482 & 0.857 & 0.840 & 0.661 &  0.607\\
Lip      & RN50     & Caltech101     & euphonium, minaret, platypus                     & 0.827 & 0.125  & 0.558 & 0.549 & 0.517 & 0.515 & --    & -- & 0.661 & 0.633   \\
Lip      & RN50     & OxfordFlowers  & gazania, tree mallow, trumpet creeper            &  0.86 & 0.0 & 0.558 & 0.552 & 0.517 & 0.498 & 0.857 & 0.863 & --    & --\\
ZSL-CLIP & RN50     & StanfordDogs   & Pekinese, toy poodle, Scotch terrier           & 0.591 & 0.0 & 0.558 & 0.540 & --    & --    & 0.857 & 0.854 & 0.661 & 0.629\\
ZSL-CLIP & RN50     & StanfordCars   & 2009 Spyker C8 Coupe, 2010 Dodge Ram Pickup 3500 Crew Cab, 2011 Ford Ranger SuperCab  & 0.397 & 0.0 & --    & --    & 0.517 & 0.499 & 0.857 & 0.850 & 0.661 & 0.654\\
ZSL-CLIP & RN50     & Caltech101     & euphonium, minaret, platypus                     & 0.827 & 0.0  & 0.558 & 0.551 & 0.517 & 0.499 & --    & -- & 0.661 & 0.655  \\
ZSL-CLIP & RN50     & OxfordFlowers  & trumpet creeper, gazania, tree mallow            & 0.86 & 0.0 & 0.558 & 0.554 & 0.517 & 0.502 & 0.857 & 0.856 & --    & --\\
Lip      & ViT-B/16 & StanfordDogs   & Pekinese, toy poodle, Scotch terrier           & 0.672  & 0.251 & 0.655 & 0.644 & --    & --    & 0.933 & 0.939 & 0.708 & 0.713\\
Lip      & ViT-B/16 & StanfordCars   & 2009 Spyker C8 Coupe, 2010 Dodge Ram Pickup 3500 Crew Cab, 2011 Ford Ranger SuperCab  & 0.595 & 0.3 & --    & --    & 0.591 & 0.576 & 0.933 & 0.928 & 0.708 &0.699\\
Lip      & ViT-B/16 & Caltech101     & euphonium, minaret, platypus                     & 0.971 & 0.498 & 0.655 & 0.634 & 0.591 & 0.589 & --    & --   & 0.708 & 0.709 \\
Lip      & ViT-B/16 & OxfordFlowers  & trumpet creeper, gazania, tree mallow            & 0.807 & 0.31 & 0.655 & 0.613 & 0.591 & 0.551 & 0.933 & 0.929  & --    & --\\
ZSL-CLIP & ViT-B/16 & StanfordDogs   & Pekinese, toy poodle, Scotch terrier           & 0.672 & 0.0 & 0.655 & 0.624 & --    & --    & 0.933 & 0.920 & 0.708 & 0.668\\
ZSL-CLIP & ViT-B/16 & StanfordCars   & 2009 Spyker C8 Coupe, 2010 Dodge Ram Pickup 3500 Crew Cab, 2011 Ford Ranger SuperCab  & 0.595 & 0.0 & --    & --    & 0.591 & 0.586 & 0.933 & 0.931 & 0.708 & 0.693\\
ZSL-CLIP & ViT-B/16 & Caltech101     & euphonium, minaret, platypus                     & 0.962 & 0.0 & 0.655 & 0.650 & 0.591 & 0.558 & --    & --  & 0.708 & 0.685  \\
ZSL-CLIP & ViT-B/16 & OxfordFlowers  & trumpet creeper, gazania, tree mallow            & 0.807 & 0.0 & 0.655 & 0.649 & 0.591 & 0.578 & 0.933 & 0.929  & --  & -- \\
\midrule
Our      & ViT-32   & StanfordDogs   & Pekinese, toy poodle, Scotch terrier           & 0.708 & 0.0 & 0.684    & 0.681    & -    & -    & 0.945    & 0.941    & 0.706    & 0.693 \\
Our      & ViT-32   & StanfordCars   & 2009 Spyker C8 Coupe, 2010 Dodge Ram Pickup 3500 Crew Cab, 2011 Ford Ranger SuperCab  & 0.609 & 0.0017 & 0.679    & 0.679    & 0.614    & 0.609    & 0.945    & 0.940    & 0.706    & 0.702 \\
Our      & ViT-32   & Caltech101     & euphonium, minaret, platypus                     & 0.958 & 0.0 & 0.668    & 0.648    & 0.614    & 0.602    & --    & --    & 0.706   & 0.701 \\
Our      & ViT-32   & OxfordFlowers  & trumpet creeper, gazania, tree mallow            & 0.842 & 0.0 & 0.679    & 0.663    & 0.614    & 0.605    & 0.945    & 0.927    & --    & -- \\
\midrule
Our      & ViT-16   & StanfordDogs   & Pekinese, toy poodle, Scotch terrier           & 0.695 & 0.0 & 0.681    & 0.668    & --    & --    & 0.938    & 0.921    & 0.702    & 0.686 \\
Our      & ViT-16   & StanfordCars   & 2009 Spyker C8 Coupe, 2010 Dodge Ram Pickup 3500 Crew Cab, 2011 Ford Ranger SuperCab  & 0.602 & 0.003 & 0.667    & 0.661    & 0.609    & 0.586    & 0.938    & 0.915    & 0.702    & 0.681 \\
Our      & ViT-16   & Caltech101     & euphonium, minaret, platypus                     & 0.958 & 0.0 & 0.673    & 0.672    & 0.609    & 0.602    & --    & --    & 0.702    & 0.685 \\
Our      & ViT-16   & OxfordFlowers  & trumpet creeper, gazania, tree mallow            & 0.821 & 0.0 & 0.670    & 0.660    & 0.609    & 0.600    & 0.938   & 0.916    & --    & -- \\
\bottomrule
\end{tabular}%
}
\label{tab:tab3}
\end{table*}

\section{Ablation Study}

To ensure that our projection unlearns targeted classes in a controlled manner while preserving the retained classes, we incorporated both an identity preservation component (C1) and a retention mechanism (C3) into our approach. An ablation study presented in Table 4 demonstrates the contribution of these components. The results show that without the identity preservation component, our projection struggles to effectively maintain the retained classes when there is overlap with the forgetting classes—this is due to the overall structure of the CLIP features not being preserved in the projected space. Similarly, omitting the retention component (C3) leads to a reduction in the accuracy of the retained classes after unlearning, and some information from the unlearned classes persists. Therefore, for effective projection learning, it is essential to include both the identity preservation and retention components.

We also examined the impact of the hyperparameters $\lambda$ and $\mu$ in our approach. To do this, we fixed $\mu$ and varied $\lambda$, evaluating the performance for each value of $\lambda$. Similarly, we fixed $\lambda$ and varied $\mu$, assessing the effects on both unlearning and retention accuracies. Table~\ref{tab:ab1} shows the results of varying $\lambda$, and Table~\ref{tab:ab2} presents the outcomes for varying $\mu$ on the StanfordDogs and StanfordCars datasets.

We observed that, with $\mu$ fixed, a smaller $\lambda$ leads to less unlearning, while increasing $\lambda$ enhances the unlearning effect. Conversely, with $\lambda$ fixed, a smaller $\mu$ results in more effective forgetting; however, as $\mu$ increases, the unlearning effect decreases. In our experiments, we set $\lambda=100$ when varying $\mu$, and fixed $\mu=1$ when varying $\lambda$.

\begin{table*}[ht]
\centering
\caption{The results demonstrate the contribution of each component in our proposed approach. Specifically, C1 corresponds to Identity Preservation, C2 represents the Forgetting Mechanism, and C3 denotes the Retention Component.}

\begin{tabular}{l | cccc | cccc | cccc}
\toprule
\multirow{3}{*}{Component} 
  & \multicolumn{4}{c|}{StanfordCars} 
  & \multicolumn{4}{c|}{StanfordDogs} 
  & \multicolumn{4}{c}{Caltech101} \\
\cmidrule(lr){2-5} \cmidrule(lr){6-9} \cmidrule(lr){10-13}
  & \multicolumn{2}{c|}{Target Class} & \multicolumn{2}{c|}{Retain Class} 
  & \multicolumn{2}{c|}{Target Class} & \multicolumn{2}{c|}{Retain Class} 
  & \multicolumn{2}{c|}{Target Class} & \multicolumn{2}{c}{Retain Class} \\
\cmidrule(lr){2-3} \cmidrule(lr){4-5} 
\cmidrule(lr){6-7} \cmidrule(lr){8-9} 
\cmidrule(lr){10-11} \cmidrule(lr){12-13}
  & BF & AF & BF & AF & BF & AF & BF & AF & BF & AF & BF & AF \\
\midrule
C1+C2 & 82.20 & 0.41 & 81.45 & 68.50 & 80.38 & 0.0 & 60.18 & 42.5 & 83.76 & 0.25 & 95.13 & 88.6 \\
C2+C3 & 82.20 & 0.52 & 81.45 & 75.1 & 80.38 & 0.04 & 60.18 & 47.6 & 83.76 & 0.85 & 95.13 & 92.7 \\
C1+C2+C3 & 82.20 & 0.38 & 81.45 & 75.42 & 80.38 & 0.0 & 60.18 & 50.30 & 83.76 & 0.0 & 95.13 & 94.44 \\
\bottomrule
\end{tabular}
\label{tab:tab4}
\end{table*}

\begin{table}[ht]
\centering
\caption{Fixed $\mu$ with increasing $\lambda$ (1, 10, 100) for two datasets.}
\resizebox{\linewidth}{!}{
\begin{tabular}{c|cc|cc|cc|cc}
\toprule
$\lambda$ & \multicolumn{4}{c|}{StanfordDogs} & \multicolumn{4}{c}{StanfordCars} \\
\cmidrule(lr){2-5} \cmidrule(lr){6-9}
      & \multicolumn{2}{c|}{Retain Set} & \multicolumn{2}{c|}{Forget Set} & \multicolumn{2}{c|}{Retain Set} & \multicolumn{2}{c}{Forget Set} \\
\cmidrule(lr){2-3} \cmidrule(lr){4-5} \cmidrule(lr){6-7} \cmidrule(lr){8-9}
      & BF & AF & BF & AF & BF & AF & BF & AF \\
\midrule
1     & 60.18 & 58.1 & 80.38 & 40.2 & 81.45 & 79.5 & 82.20 & 38.8 \\
10    & 60.18 & 58.3 & 80.38 & 29.6& 81.45 & 79.5 & 82.20 & 25.7 \\
100   & 60.18 & 58.9 & 80.38 & 4.8 & 81.45 & 79.8 & 82.20 & 7.5 \\
\bottomrule
\end{tabular}
}
\label{tab:ab1}
\end{table}

\begin{table}[ht]
\centering
\caption{Fixed $\lambda$ with increasing $\mu$ (1, 10, 100) for two datasets.}
\resizebox{\linewidth}{!}{
\begin{tabular}{c|cc|cc|cc|cc}
\toprule
$\mu$ & \multicolumn{4}{c|}{StanfordDogs} & \multicolumn{4}{c}{StanfordCars} \\
\cmidrule(lr){2-5} \cmidrule(lr){6-9}
      & \multicolumn{2}{c|}{Retain Set} & \multicolumn{2}{c|}{Forget Set} & \multicolumn{2}{c|}{Retain Set} & \multicolumn{2}{c}{Forget Set} \\
\cmidrule(lr){2-3} \cmidrule(lr){4-5} \cmidrule(lr){6-7} \cmidrule(lr){8-9}
      & BF & AF & BF & AF & BF & AF & BF & AF \\
\midrule
1     & 60.18 & 58.9 & 80.38 & 7.60 & 81.45 & 80.6 & 82.20 & 5.3 \\
10    & 60.18 & 56.7 & 80.38 & 15.6 & 81.45 & 79.4 & 82.20 & 10.5 \\
100   & 60.18 & 57.9 & 80.38 & 18.3& 81.45 & 78.5 & 82.20 & 14.8 \\
\bottomrule
\end{tabular}
}
\label{tab:ab2}
\end{table}

\begin{figure}[h]
\centering
\includegraphics[width=0.5\textwidth]{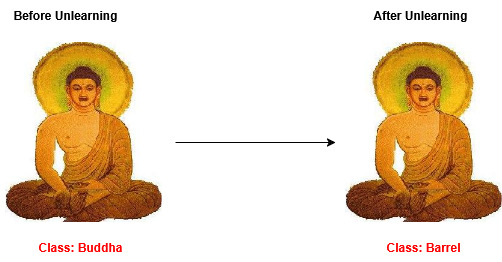}
\caption{Class prediction in before and after unlearning.}
\label{fig:ab1}
\end{figure}

We also present visualization results comparing class predictions before and after class unlearning, using both the CLIP-projected features and our proposed projection features. In Figure \ref{fig:ab1}, the model correctly predicts the class as Buddha before unlearning; however, after removing the knowledge of the Buddha class, it incorrectly predicts the class as Barrel.

\textbf{Computation complexity:} The main computational cost of our proposed approach lies in calculating the projection matrix, after which it provides a closed-form solution. In contrast, most baseline methods require fine-tuning or additional training on generated images of the forgotten classes. As a result, our method is significantly more computationally efficient than the baselines.

\section{Conclusion}
\label{sec:conclusion}
In this paper, we proposed a framework to unlearn the \textit{forget class} information from the CLIP model while preserving the \textit{retain class} information. Our approach, Consistent Class Unlearning Projection (CCUP), introduces a simple yet effective projection technique that effectively suppresses the features of the forget class while maintaining the integrity of the retain class. Experimental results demonstrate that CCUP consistently outperforms baseline methods, achieving superior unlearning while minimizing unintended degradation in retained knowledge.

Our findings open up exciting possibilities for tackling a wide range of challenges in machine unlearning, fairness, and privacy preservation. The ability to selectively unlearn specific information while maintaining overall model performance is crucial for applications in data compliance (e.g., GDPR "right to be forgotten"), bias mitigation, and model refinement in dynamic environments. Future work can explore extending CCUP to more complex multimodal settings and larger-scale datasets, ensuring its adaptability to real-world scenarios. Additionally, integrating CCUP with continual learning frameworks could allow models to evolve over time while still adhering to ethical and regulatory constraints.

Another promising direction is the development of adaptive unlearning mechanisms that adjust dynamically based on model confidence and user-defined criteria, improving efficiency and robustness. Moreover, exploring CCUP’s applicability in adversarial robustness, misinformation mitigation, and personalized AI systems could unlock new frontiers in trustworthy AI development. Ultimately, our work lays a strong foundation for future advancements in selective forgetting, guiding the design of more responsible and adaptable AI models.

\clearpage
{
    \small
    \bibliographystyle{ieeenat_fullname}
    \bibliography{main}
}

\end{document}